\documentclass[conference]{IEEEtran}
\IEEEoverridecommandlockouts
\usepackage{cite}
\usepackage{amsmath,amssymb,amsfonts}
\usepackage{algorithmic}
\usepackage{graphicx}
\usepackage{textcomp}
\usepackage{xcolor}
\usepackage[english]{babel}
\usepackage{color,soul}
\usepackage{csquotes}
 \usepackage{array,multirow,graphicx}
 \usepackage{float}
\usepackage{array, makecell} %
\usepackage{tabularx,booktabs}
\usepackage{graphicx}
\usepackage{fancyhdr}
\usepackage{amsmath}
\usepackage{listings} %if lstlistings is used
\usepackage{changepage} %for changing topmargin on first page
\usepackage[figurename=Fig., tablename=Tab., small]{caption}%[2008/04/01]
\usepackage{color}
\usepackage{xcolor}
\usepackage{footmisc}
\usepackage{url}
\usepackage{booktabs}
\usepackage{times}
\usepackage{verbatim}

\def\BibTeX{{\rm B\kern-.05em{\sc i\kern-.025em b}\kern-.08em
    T\kern-.1667em\lower.7ex\hbox{E}\kern-.125emX}}
\begin{document}

\title{OrthoMAD: Morphing Attack Detection Through Orthogonal Identity Disentanglement
\thanks{
This work was financed by National Funds through the Portuguese funding agency, FCT - Fundação para a Ciência e a Tecnologia within project LA/P/0063/2020 and the PhD grants ‘‘2021.06872.BD’’ and ‘‘UIDB/50014/2020’’. This research work has been also funded by the German Federal Ministry of Education and Research and the Hessen State Ministry for Higher Education, Research and the Arts within their joint support of the National Research Center for Applied Cybersecurity ATHENE.}
}

\author{\IEEEauthorblockN{Pedro C. Neto\textsuperscript{1,2}, Tiago Gonçalves\textsuperscript{1,2}, Marco Huber\textsuperscript{3,4}, Naser Damer\textsuperscript{3,4},\\Ana F. Sequeira\textsuperscript{1}, and Jaime S. Cardoso\textsuperscript{1,2}}
\IEEEauthorblockA{\textsuperscript{1}\textit{Centre for Telecommunications and Multimedia, INESC TEC, Porto, Portugal} \\
\textsuperscript{2}\textit{Faculdade de Engenharia da Universidade do Porto, Porto, Portugal}\\
\textsuperscript{3}\textit{Fraunhofer Institute for Computer Graphics Research (IGD), Darmstadt, Germany}\\
\textsuperscript{4}\textit{Technische Universität Darmstadt, Darmstadt, Germany}\\
pedro.d.carneiro@inesctec.pt
}
}

\maketitle

\begin{abstract}
Morphing attacks are one of the many threats that are constantly affecting deep face recognition systems. It consists of selecting two faces from different individuals and fusing them into a final image that contains the identity information of both. In this work, we propose a novel regularisation term that takes into account the existent identity information in both and promotes the creation of two orthogonal latent vectors. We evaluate our proposed method  (OrthoMAD) in five different types of morphing in the FRLL dataset and evaluate the performance of our model when trained on five distinct datasets.  With a small ResNet-18 as the backbone, we achieve state-of-the-art results in the majority of the experiments, and competitive results in the others.
\end{abstract}

\section{Introduction}

Over the years, the performance of face recognition systems kept increasing significantly. Larger datasets have been collected, and better models have been developed and published~\cite{grother2018ongoing}. Hence, its usage has been proportional to its performance, leading to a wide spread of these algorithms. However, in the real world, these models might face attacks that aim to increase the number of false positives given by the model~\cite{kramer2019face}. These attacks vary significantly. For instance, one may add a simple facial mask to a user's face or print the target user's face and use it as a presentation attack to the system. Hence, researchers also focus on developing detection approaches to these threats \cite {neto2022myope,neto2021focusface}. The former does not care about which identity the model identifies, whereas the latter aims to create a positive sample based on the presented identity.

Another category of attacks, known as morphing attacks, aims to incorporate facial features from two distinct identities in a single image crafted from a fusion of the images corresponding to these two identities~\cite{scherhag2019face}. Hence, one may use this newly designed image to allow two distinct users to enter the system or two different people to use the same passport and pass border control. Besides, similarly to the research on morphing attacks, the community has also dedicated several research efforts toward detecting these attacks~\cite{damer2018morgan,damer2021pw,tapia2021single,huberCompetition}. Usually, these methods do not use any information regarding the identity fused in the attacks and focus the model's training on detecting a potential attack. Therefore, it is often common to formulate these problems as binary classification tasks where the negative label is associated with an attack and the positive with a \textit{bona fide} sample.

The work proposed in this paper includes a new regularisation term to a modified ResNet-18~\cite{he2016deep} architecture trained to detect morphing attacks. Besides the binary classification system, this new framework presents two vectors in different latent spaces. These vectors, known as identity vectors, are strongly regularised to be orthogonal, thus, promoting independent identity information in each. Moreover, although we used ResNet-18 in our experiments, one may apply this methodology on top of any existing architecture.

The main contributions of this work are: 1) the addition of a loss regularisation term that aims to promote orthogonality between the identity-related embeddings (i.e. latent space vectors); 2) the empirical validation of the proposed loss in a large set of publicly available datasets of face morphing attacks; 3) a comparison with the state-of-the-art approaches presented in the literature that use the same dataset.

\begin{figure*}[h!]
    \centering
    \includegraphics[width=\linewidth]{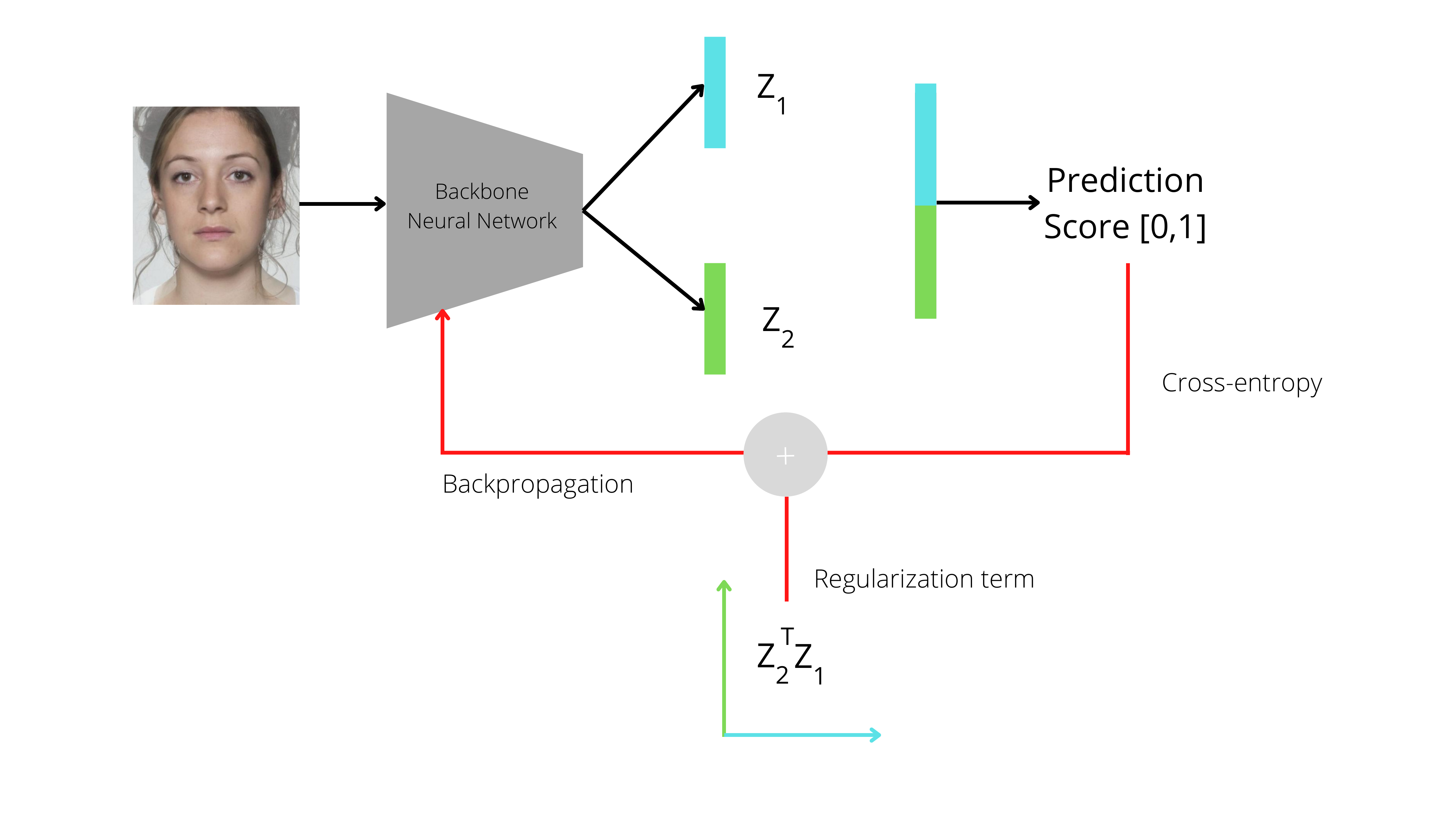}
    \caption{Overview of the architecture design for the integration of the new regularisation term. The regularisation loss is applied at the vector ($Z_1$ and $Z_2$), hence, their gradients do not affect the final classification layer. The system is composed of three components, a backbone network, two identity vectors and a final classification layer that uses the concatenation of both vectors. }
    \label{fig:my_label}
\end{figure*}

In this Introduction, we discriminated the types of attacks that affect biometric systems and which we are trying to detect and enumerated our main contributions. Moreover, we organised the remainder of this paper as follows: Section~\ref{sec:datasets} describes in detail the datasets used to design the experiments; Section~\ref{sec:methods} presents the architecture design of the proposed regularisation approach; Section~\ref{sec:experiments} reports and discuss the performance of our method in the selected datasets; and Section~\ref{sec:conclusion}, concludes this paper with some final thoughts and future work suggestions. The code related to this paper is publicly available in a GitHub repository\footnote[1]{\url{https://github.com/NetoPedro/OrthoMAD}}.

\section{Databases}\label{sec:datasets}

\paragraph{MorGAN}
Divided into two sets, the MorGAN dataset~\cite{damer2018morgan} fused images from the CelebA~\cite{liu2015faceattributes} dataset as morphing attacks. Images were selected based on their frontal pose, and the OpenFace~\cite{baltruvsaitis2016openface} FR solution was used to find the most similar pairs to morph~\cite{liu2015deep}. The morphing approach followed relied on the landmark-based OpenCV~\cite{Mallick_2016} or a GAN-based~\cite{damer2018morgan} solution.  These different approaches resulted in the two distinct sets MorGAN-LMA and MorGAN-GAN, respectively. Each of the train and test sets (available in both sets) contains 500 morphing attacks and 750 \textit{bona fide} samples. This dataset contains images of low resolution (64x64).

\paragraph{FRLL} Extensively used to evaluate morphing attack detection algorithms, the FRLL-Morphs dataset~\cite{sarkar2020vulnerability} was created from the publicly available Face Research London Lab dataset~\cite{debruine_jones_2021}. The dataset contains five distinct morphing approaches, including Style-GAN2~\cite{karras2020training,venkatesh2020can}, WebMorph~\cite{debruine2018debruine}, AMSL~\cite{neubert2018extended}, FaceMorpher~\cite{quek2019facemorpher} and OpenCV~\cite{Mallick_2016}. Each of the five approaches contains 1222 morphed faces created from frontal faces with high resolution and 204 \textit{bona fide} samples. This database does not contain disjoint train/test sets, hence it was used just for evaluation.

\paragraph{LMA-DRD}
The LMA-DRD dataset~\cite{damer2021pw} uses the images published with the VGGFace2~\cite{cao2018vggface2} dataset to create the morphing attack. Despite having more than 3 million images available, not all were selected to be fused into attacks. The selection requirements included a frontal pose, high resolution, and a neutral expression. Using the parametrisation introduced in~\cite{raghavendra2017face}, the images were morphed with the OpenCV morphing~\cite{Mallick_2016}. \textit{Bona fide} images were selected using the previously mentioned criteria. This dataset has two versions, a digital - LMA-DRD (D) - and a printed and scanned - LMA-DRD (PS). We used the train set for training (96 morphs and 121 BF images), and the identity-disjoint test set for testing (88 morphs and 123 \textit{bona fide} (BF) images).

\paragraph{SMDD} The Synthetic Morphing Attack Detection Development (SMDD)~\cite{damer2022privacy} utilised the official open-source implementation of StyleGan2-ADA~\cite{karras2020training} to generate 500k images of faces using a random Gaussian noise vector sampled from a normal distribution. From these images, 50k were selected due to their high quality (using CR-FIQA~\cite{boutros2021cr}), and 25k from those were considered \textit{bona fide}. 5k of the non-\textit{bona fide} images were selected as key morphing images and paired with five randomly chosen non-\textit{bona fide} images. They were then morphed with the OpenCV~\cite{Mallick_2016} approach, generating 15k attack samples.

\begin{table*}[ht!]
 \caption{Results for the different types of morphing techniques included in the FRLL dataset. For morphing type, we present the performance of the model per training dataset. All the results are in percentage (\%), and the best are in bold.}
\label{res-rmfd}
\centering
\resizebox{0.68\textwidth}{!}{
\begin{tabular}{|c|l|c|c|c|c|}
\hline
& & & & \multicolumn{2}{c|}{BPCER @ APCER =} \\
\cline{1-6}
Test&Train      & Model & EER &  1\% &  20\% \\ 
\cline{1-6}
 \parbox[t]{0mm}{\multirow{5}{*}{\rotatebox[origin=c]{90}{FRLL - Style-GAN2}}}&MorGAN-LMA & \makecell{ResNet-18\\\textbf{OrthoMAD (Ours)}}& \makecell{\textbf{28.72}\\51.47} & \makecell{99.01\\\textbf{97.46}}  &  \makecell{\textbf{60.81}\\81.51}      \\
 \cline{2-6}
 &MorGAN-GAN &\makecell{ResNet-18\\\textbf{OrthoMAD (Ours)}}& \makecell{53.60\\\textbf{17.34}} & \makecell{99.59\\\textbf{59.81}}  &  \makecell{87.07\\\textbf{14.32}}     \\
 \cline{2-6}
  &LMA-DRD (D) &\makecell{ResNet-18\\\textbf{OrthoMAD (Ours)}}& \makecell{62.60\\\textbf{31.50}} & \makecell{100.00\\\textbf{99.91}}  &  \makecell{97.87\\\textbf{63.74}}     \\
 \cline{2-6}
  &LMA-DRD (PS)& \makecell{ResNet-18\\\textbf{OrthoMAD (Ours)}} & \makecell{71.19\\\textbf{29.21}} & \makecell{100.00\\\textbf{99.59}}  &  \makecell{97.38\\\textbf{53.02}}      \\
 \cline{2-6}
  &SMDD & \makecell{ResNet-18\\\textbf{OrthoMAD (Ours)}} & \makecell{11.62\\\textbf{6.54}} & \makecell{35.18\\\textbf{13.74}}  &  \makecell{7.77\\\textbf{3.76}}  \\
\cline{1-6}

 \parbox[t]{0mm}{\multirow{5}{*}{\rotatebox[origin=c]{90}{FRLL - WebMorph}}}&MorGAN-LMA &\makecell{ResNet-18\\\textbf{OrthoMAD (Ours)}} & \makecell{37.10\\\textbf{8.35}} & \makecell{99.09\\\textbf{45.61}}  &  \makecell{60.68\\\textbf{2.53}}  \\
 \cline{2-6}
 &MorGAN-GAN &\makecell{ResNet-18\\\textbf{OrthoMAD (Ours)}}& \makecell{23.25\\\textbf{14.41}} & \makecell{89.59\\\textbf{45.12}}  &  \makecell{25.71\\\textbf{9.90}}     \\
 \cline{2-6}
  &LMA-DRD (D) &\makecell{ResNet-18\\\textbf{OrthoMAD (Ours)}} & \makecell{93.85\\\textbf{23.58}} & \makecell{100.00\\\textbf{97.37}}  &  \makecell{100.00\\\textbf{29.40}}  \\
 \cline{2-6}
  &LMA-DRD (PS)& \makecell{ResNet-18\\\textbf{OrthoMAD (Ours)}} & \makecell{88.62\\\textbf{10.65}} & \makecell{100.00\\\textbf{60.28}}  &  \makecell{100.00\\\textbf{7.20}}  \\
 \cline{2-6}
  &SMDD & \makecell{ResNet-18\\\textbf{OrthoMAD (Ours)}} & \makecell{16.29\\\textbf{15.23}} & \makecell{85.33\\\textbf{70.92}}  &  \makecell{16.29\\\textbf{9.50}}  \\
\cline{1-6}

 \parbox[t]{0mm}{\multirow{5}{*}{\rotatebox[origin=c]{90}{FRLL - AMSL}}}&MorGAN-LMA &\makecell{ResNet-18\\\textbf{OrthoMAD (Ours)}} & \makecell{54.16\\\textbf{0.91}} & \makecell{99.63\\\textbf{0.91}}  &  \makecell{84.36\\\textbf{0.00}}  \\
 \cline{2-6}
  &MorGAN-GAN&\makecell{ResNet-18\\\textbf{OrthoMAD (Ours)}} & \makecell{44.59\\\textbf{7.91}} & \makecell{\textbf{90.94}\\97.01}  &  \makecell{63.72\\\textbf{1.19}}  \\
 \cline{2-6}
 &LMA-DRD (D) &\makecell{ResNet-18\\\textbf{OrthoMAD (Ours)}}& \makecell{50.67\\\textbf{30.43}} & \makecell{95.86\\\textbf{89.70}}  &  \makecell{75.86\\40.46}     \\
 \cline{2-6}
  &LMA-DRD (PS)& \makecell{ResNet-18\\\textbf{OrthoMAD (Ours)}} & \makecell{83.35\\\textbf{27.26}} & \makecell{100.00\\\textbf{87.21}}  &  \makecell{100.00\\\textbf{37.19}}  \\
 \cline{2-6}
  &SMDD & \makecell{ResNet-18\\\textbf{OrthoMAD (Ours)}} & \makecell{34.43\\\textbf{14.80}} & \makecell{65.97\\\textbf{65.05}}  &  \makecell{27.26\\\textbf{10.89}}  \\
\cline{1-6}
 \parbox[t]{0mm}{\multirow{5}{*}{\rotatebox[origin=c]{90}{FRLL - FaceMorpher}}}&MorGAN-LMA &\makecell{ResNet-18\\\textbf{OrthoMAD (Ours)}}& \makecell{\textbf{5.48}\\34.20} & \makecell{\textbf{25.45}\\97.29}  &  \makecell{\textbf{1.23}\\54.90}  \\
 \cline{2-6}
  &MorGAN-GAN&\makecell{ResNet-18\\\textbf{OrthoMAD (Ours)}}& \makecell{37.15\\\textbf{35.27}} & \makecell{97.95\\\textbf{94.10}}  &  \makecell{\textbf{55.72}\\56.71}  \\
 \cline{2-6}
  &LMA-DRD (D)& \makecell{ResNet-18\\\textbf{OrthoMAD (Ours)}} & \makecell{39.60\\\textbf{30.19}} & \makecell{99.59\\\textbf{83.87}}  &  \makecell{76.51\\\textbf{38.13}}  \\
 \cline{2-6}
 &LMA-DRD (PS)& \makecell{ResNet-18\\\textbf{OrthoMAD (Ours)}} & \makecell{40.02\\\textbf{34.12}} & \makecell{\textbf{99.42}\\99.83}  &  \makecell{\textbf{70.21}\\71.84}  \\
 \cline{2-6}
  &SMDD & \makecell{ResNet-18\\\textbf{OrthoMAD (Ours)}} & \makecell{2.95\\\textbf{0.98}} & \makecell{5.32\\\textbf{2.37}}& \makecell{2.37\\\textbf{0.08}}\\
\cline{1-6}

\parbox[t]{0mm}{\multirow{5}{*}{\rotatebox[origin=c]{90}{FRLL - OpenCV}}}&MorGAN-LMA &\makecell{ResNet-18\\\textbf{OrthoMAD (Ours)}}& \makecell{\textbf{16.29}\\27.92} & \makecell{\textbf{64.61}\\93.44}  &  \makecell{\textbf{11.62}\\38.82}  \\
 \cline{2-6}
  &MorGAN-GAN&\makecell{ResNet-18\\\textbf{OrthoMAD (Ours)}}& \makecell{40.29\\\textbf{29.07}} & \makecell{99.34\\\textbf{97.21}}  &  \makecell{79.11\\\textbf{43.48}}  \\
 \cline{2-6}
  &LMA-DRD (D)& \makecell{ResNet-18\\\textbf{OrthoMAD (Ours)}} & \makecell{66.75\\\textbf{33.98}} & \makecell{99.83\\\textbf{99.59}}  &  \makecell{96.56\\\textbf{60.68}}  \\
 \cline{2-6}
 &LMA-DRD (PS)& \makecell{ResNet-18\\\textbf{OrthoMAD (Ours)}} & \makecell{49.63\\\textbf{42.17}} & \makecell{100.00\\\textbf{99.09}}  &  \makecell{85.17\\\textbf{73.87}}  \\
 \cline{2-6}
  &SMDD & \makecell{ResNet-18\\\textbf{OrthoMAD (Ours)}} & \makecell{1.22\\\textbf{0.73}} & \makecell{11.8\\\textbf{0.73}}& \makecell{0.41\\\textbf{0.32}}\\
\cline{1-6}

\end{tabular}
}
\end{table*}

% Methodology

\begin{table*}[ht!]
 \caption{Results comparison with three other models published in the literature. All the models were trained on the SMDD dataset. All the results are in percentage (\%) and the best are in bold. }
\label{res-rmfd2}
\centering
\resizebox{0.6\textwidth}{!}{
\begin{tabular}{|c|c|c|c|c|}
\hline
& &  & \multicolumn{2}{c|}{BPCER @ APCER =} \\
\cline{1-5}
Test& Model & EER &  1\% &  20\% \\ 
\cline{1-5}
FRLL-Style-GAN2 & \makecell{Inception\\PW-MAD\\MixFacenet\\\textbf{OrthoMAD (Ours)}}& \makecell{11.37\\16.64 \\8.99 \\\textbf{6.54}} & \makecell{72.06\\80.39 \\42.16 \\\textbf{13.74}}  &  \makecell{6.86\\13.24 \\4.41 \\\textbf{3.76}}   \\
 \cline{1-5}
FRLL-WebMorph & \makecell{Inception\\PW-MAD\\MixFacenet\\\textbf{OrthoMAD (Ours)}}& \makecell{\textbf{9.86}\\16.65 \\12.35 \\15.23} &   \makecell{\textbf{53.92}\\80.39 \\80.39 \\70.92}  &  \makecell{\textbf{2.94}\\13.24 \\7.84 \\9.50} \\
 \cline{1-5}
FRLL-OpenCV&\makecell{Inception\\PW-MAD\\MixFacenet\\\textbf{OrthoMAD (Ours)}} & \makecell{5.38\\2.42 \\4.39 \\\textbf{0.73}} & \makecell{38.73\\22.06 \\26.47 \\\textbf{0.73}} &  \makecell{0.98\\0.49 \\1.47 \\\textbf{0.32}} \\
 \cline{1-5}
 FRLL-AMSL& \makecell{Inception\\PW-MAD\\MixFacenet\\\textbf{OrthoMAD (Ours)}} & \makecell{\textbf{10.79}\\15.18 \\15.18 \\14.80} & \makecell{72.06\\96.57 \\\textbf{49.51} \\65.05} & \makecell{\textbf{4.90}\\5.88 \\11.76 \\10.89}\\
\cline{1-5}
 FRLL-FaceMorpher  & \makecell{Inception\\PW-MAD\\MixFacenet\\\textbf{OrthoMAD (Ours)}} & \makecell{3.17\\ 2.20\\3.87 \\\textbf{0.98}} & \makecell{30.39\\ 26.47\\23.53 \\\textbf{2.37}} & \makecell{0.49\\ \textbf{0.00}\\0.49 \\0.08} \\
\cline{1-5}

\end{tabular}
}
\end{table*}

\section{Methodology}\label{sec:methods}
Morphing attacks result from a fusion process of two distinct identities. In the final image, there is enough information about both identities to trick a face recognition system. Hence, to leverage the presence of such information, we designed a regularisation term to promote the separation of the information from both identities. When two vectors are orthogonal, there is no common information in them. The orthogonality of these two vectors is then the desired property when trying to disentangle identity information into separate vectors. And thus, we introduce a regularisation term (Eq.~\ref{eq:inner}), which leverages the inner product of two vectors.

\begin{align}
\label{eq:inner}
    Reg = (Z_1^T Z_2)^2
\end{align}

Its integration in the final loss is straightforward since the main objective is to minimise the inner product to zero. We further square the inner product to make the optimisation smoother. To be able to have these two vectors, the architecture of a ResNet-18 had to be adapted. The last fully-connected layer was replaced with two fully-connected layers that output a vector of size 32 each. These vectors are used to apply the orthogonal regularisation and are afterwards concatenated (Eq.~\ref{eq:concat}).

\begin{equation}
\label{eq:concat}
    Z = concat(Z_1, Z_2)
\end{equation}

As seen in Fig.~\ref{fig:my_label}, this concatenation is directly fed into another fully connected layer to produce a score $Y$, which is activated with the sigmoid non-linearity (Eq.~\ref{eq:sigmoid}). 

\begin{equation}
\label{eq:sigmoid}
    Y = \frac{1} {1 + e^{-(W^TZ)}}
\end{equation}

The loss of the prediction is calculated using the Binary Cross-Entropy (Eq.~\ref{eq:bce}), and the regularisation term is summed to the final loss with a certain weight controlled by the hyperparameter $\alpha$ (Eq.~\ref{eq:loss}). An illustration of the end-to-end optimisation and classification process is seen in Fig.~\ref{fig:my_label}.

\begin{equation}
\label{eq:bce}
    \textup{\textit{Binary Cross Entropy}} = -{(y\log(p) + (1 - y)\log(1 - p))}
\end{equation}

\begin{equation}
\label{eq:loss}
    Loss = -{(y\log(p) + (1 - y)\log(1 - p))} + \alpha (Z_2^T Z_1)^2
\end{equation}

\paragraph{Experimental setup} We design the experiments to evaluate the performance of our approach when compared to a baseline architecture that does not include regularisation. For this, we train both models in all the datasets and evaluate them in the five different types of morphing approaches found in the FRLL dataset. We trained all the models with a batch size of 16, a learning rate of $10^{-5}$, and an $\alpha$ hyperparameter of 100. We further augmented the dataset with random horizontal flips and resized all the images to be $224\times224$. Finally, we also crop the faces from the original images. The detection performance is shown by the Attack Presentation Classification Error Rate (APCER) (i.e., attacks classified as \textit{bona fide}), and the \textit{Bona fide} Presentation Classification Error Rate (BPCER) (i.e., the \textit{bona fide} samples classified as attacks). We report the BPCER at two different fixed APCER values (1.0\% and 20.0\%). We also report the equal error rate (EER), which is the BPCER and APCER at the decision thresholds where they are the same.

\section{Results and Discussion}\label{sec:experiments}
In this Section, we discuss the results obtained with our regularisation term (OrthoMAD) and compare it with a version without the regularisation. The performance of OrthoMAD indicates an EER lower than the EER for ResNet-18 in 22 out of 25 evaluations (see Tab.~\ref{res-rmfd}). Nonetheless, the values of BPCER @ APCER \{1\%,20\%\} seem to vary more when the EER is close in both models. While the performance of OrthoMAD indicates good performance and generalisation capabilities across datasets, it is worth noting that it has some difficulties when handling certain types of morphing (e.g., FaceMorpher) when the training dataset is small. Hence, this model excels in the majority of the testing datasets when trained in a large-scale training set (SMDD), even when the dataset is solely composed of synthetic images, which might hold less ``identity'' information.

We further extend our comparison to include the models published in the literature (see Tab.~\ref{res-rmfd2}). Hence, we compare with the results shown by Damer~\textit{et al.}~\cite{damer2022privacy} on the Inception~\cite{szegedy2016rethinking}, the PW-MAD~\cite{damer2021pw} and the MixFacenet~\cite{boutros2021mixfacenets} models. OrthoMAD outperforms all these methods in three out of five evaluation settings and is competitive in the remaining two. These results show that even with a less parameterised backbone network, our regularisation term was capable of achieving state-of-the-art performance in the task of morphing attack detection.

% COnclusions and Future Work
\section{Conclusions and Future Work}\label{sec:conclusion}
This paper presented a novel face-morphing attack detection system that promotes the disentanglement of identity features. We proposed to enforce orthogonality between the learned features of two identities in an input image, using a novel term in the loss function (i.e., regularisation). We compared our model against the state-of-the-art and obtained similar results. Nevertheless, we argue that this novel term in the loss function imposes a constraint that contributes toward the transparency of our model. With this regularisation method, we are sure that the user knows \textit{a priori} one of the rules the model must follow to output a decision. Further work should be devoted to testing different backbone architectures (e.g., attention-based models) and to the generation of saliency map-based explanations to assess if the imposition of our constraints has some significant impact on post-hoc explanation methods.

\bibliographystyle{IEEEtran}
\bibliography{refs}

\end{document}